\documentclass[journal]{IEEEtran}

\usepackage{booktabs}
\usepackage{graphicx} 
\usepackage{caption}
\usepackage{amsmath}
\usepackage{cite}
\usepackage{color}
\usepackage{subcaption}
\usepackage{float}
\usepackage{multirow}
\usepackage{booktabs}
\usepackage{amsfonts}
\newcommand{\nathaniel}[1]{{\leavevmode\color{black}{#1}}}

\ifCLASSINFOpdf
\else
\fi
\begin{document}
%
\title{
S$^{\text{4}}$TP: Social-Suitable and Safety-Sensitive Trajectory Planning for Autonomous Vehicles

}

\author{
Xiao Wang, \textit{Senior Member}, \textit{IEEE}, Ke Tang, Xingyuan Dai, Jintao Xu, \\
Quancheng Du, Rui Ai, Yuxiao Wang and Weihao Gu

\thanks{
This work was supported by the National Natural Science Foundation of China under Grant 62173329. (\textit{Corresponding author: Xiao Wang}.)

Xiao Wang is with Engineering Research Center of Autonomous Unmanned System Technology, Ministry of Education, Anhui University, Hefei 230031, China (E-mail: xiao.wang@ahu.edu.cn)

Ke Tang, Jintao Xu, Rui Ai, and Weihao Gu are with Haomo Technology Co., Ltd., Beijing 100192, China.

Xingyuan Dai is with the State Key Laboratory for Management and Control of Complex Systems, Institute of Automation, Chinese Academy of Sciences, Beijing, 100190, China, and also with Qingdao Academy of Intelligent Industries, Qingdao 266114, China.

Quancheng Du is with the School of Computer and Communication Engineering, University of Science and Technology Beijing, Beijing, 100083, China.

Yuxiao Wang is with the School of Artificial Intelligence, University of Chinese Academy of Sciences, Beijing 100049, China, and also with the State Key Laboratory for Management and Control of Complex Systems, Institute of Automation, Chinese Academy of Sciences, Beijing 100190, China.
}
}

\markboth{Journal of \LaTeX\ Class Files,~Vol.~14, No.~8, August~2023}%
{Shell \MakeLowercase{\textit{et al.}}: Bare Demo of IEEEtran.cls for IEEE Journals}

\maketitle

\begin{abstract}

In public roads, autonomous vehicles (AVs) face the challenge of frequent interactions with human-driven vehicles (HDVs), which render uncertain driving behavior due to varying social characteristics among humans.  
To effectively assess the risks prevailing in the vicinity of AVs in social interactive traffic scenarios and achieve safe autonomous driving, this article proposes a social-suitable and safety-sensitive trajectory planning (S$^{\text{4}}$TP) framework.
\nathaniel{Specifically, S$^{\text{4}}$TP integrates the Social-Aware Trajectory Prediction (SATP) and Social-Aware Driving Risk Field (SADRF) modules. SATP utilizes Transformers to effectively encode the driving scene and incorporates an AV's planned trajectory during the prediction decoding process. 
SADRF assesses the expected surrounding risk degrees during AVs-HDVs interactions, each with different social characteristics, visualized as two-dimensional heat maps centered on the AV. 
SADRF models the driving intentions of the surrounding HDVs and predicts trajectories based on the representation of vehicular interactions.
S$^{\text{4}}$TP employs an optimization-based approach for motion planning, utilizing the predicted HDVs' trajectories as input. With the integration of SADRF, S$^{\text{4}}$TP executes real-time online optimization of the planned trajectory of AV within low-risk regions, thus improving the safety and the interpretability of the planned trajectory.} 
We have conducted comprehensive tests of the proposed method using the SMARTS simulator. Experimental results in complex social scenarios, such as unprotected left-turn intersections, merging, cruising, and overtaking, validate the superiority of our proposed S$^{\text{4}}$TP in terms of safety and rationality. 
S$^{\text{4}}$TP achieves a pass rate of 100\% across all scenarios, surpassing the current state-of-the-art methods Fanta of 98.25\% and Predictive-Decision of 94.75\%.

\end{abstract}


\IEEEpeerreviewmaketitle

\section{Introduction}

Autonomous driving technologies have improved by leaps and bounds in recent years\cite{teng2023motion}. The popularization of autonomous driving will definitely bring great changes and challenges to the current transportation systems \cite{AD1,AD2,AD3,AD4}. In particular, we will step into human-driven vehicles (HDVs) and autonomous vehicles (AVs) hybrid and mixed traffic age. Therefore, considering human driving preferences, social interactions, and other social characteristics will be crucial for AVs to be aware of their situation in complex social interactive scenarios\cite{9729517}. 
In order to make safe and efficient decisions and plans, AVs need to be able to recognize and understand the differences in social preferences of various drivers and make socially appropriate decisions and plans based on these differences.

Research on the socially appropriate abilities of AVs has drawn the great attention of scholars from both academia and industry. Existing research has placed a special emphasis on decision-making methods based on social learning\cite{li2022combining, teng2022hierarchical,9815528}, socially balanced trajectory planning\cite{wang2023safety}, driving style modeling and prediction\cite{zhang2022learning,toghi2022social,S3-0}, and environmental risk assessment\cite{du2022learning, li2022pomdp}. 
However, these methods often stereotype HDVs and focus on single-traffic scenarios.
This approach compromises the adaptability across varied scenarios and the accurate representation of driver behavior, resulting in inadequate risk assessments.
Therefore, it is necessary to take social interactions into account when it comes to practical applications. 
To this end, this article focuses on decision-making and trajectory planning for AVs that consider the social suitability of the planned trajectories in complex scenarios, with the goal of achieving safer and more flexible autonomous driving levels.
 
In complex traffic scenarios with significant social interactions, such as unprotected left-turn intersections and overtaking, AVs need comprehensive sensing in all directions to make safe and efficient decisions \cite{zhang2021vehicle}. Generally, human drivers of different ages, genders, and backgrounds have varying preferences in controversial situations. 
Therefore, it is important for AVs to prioritize their capability to interact with and identify the driving styles and intentions of others, especially those encountered along their intended paths.
Then, in the absence of direct communication, AVs need to infer the intentions of the oncoming vehicles and make a risk assessment.
Finally, based on the prediction of intentions and risk assessments of other vehicles, AVs make decisions to implement trajectory planning that balances social suitability, safety, and comfort.

Modeling the surrounding scene and conducting risk assessments are the most crucial and challenging tasks for AVs in complex social interaction scenarios.
If scene modeling and risk assessment are inaccurate, AVs are likely to perform emergency braking behavior in the final decision-making stage, which reduces the comfort of autonomous driving and even causes severe accidents.
Therefore, social interactions between vehicles and the driving intentions of HDVs are important for AVs to assess the surrounding environment and achieve safe, comfortable, and efficient trajectory planning, especially in complex traffic scenarios involving multi-vehicle participation.

Currently, there are two types of risk-based trajectory planning methods: those based on trajectory prediction \cite{hubmann2018automated,wu2020trajectory} and those based on driving risk fields (DRF) 
 \cite{kim2017collision,19,du2022learning}, separately. 
The first kind of method evaluates the driving safety risks of the multimodal trajectory of AVs by predicting the trajectories of other vehicles and selecting the trajectory with the highest safety score for execution. It models and evaluates risks through the predicted trajectories.
However, this risk modeling approach is relatively coarse, making it a great challenge to achieve accurate and comprehensive assessments of the risks surrounding AVs.
Trajectory planning based on the DRF maps the surrounding risk factors of the AVs, such as HDVs, pedestrians, and obstacles, onto the DRF represented by a two-dimensional heatmap and searches for the optimal trajectory within the risk field. This method can intuitively depict the distribution of risks in the environment. 
However, most of the research on DRF currently faces two challenges. 
First, some methods concentrate solely on static risks of the immediate moment and overlook short-term future traffic conditions. 
Second, some approaches neglect the social interactions inherent between AVs and HDVs.
These limitations decrease the precision of risk assessments for AVs, making it difficult to perform reasonable trajectory planning while ensuring social suitability, safety, and efficiency.

This article proposes a social-suitable and safety-sensitive trajectory planning (S$^{\text{4}}$TP) framework for AVs, which performs trajectory planning based on a social-aware driving risk file (SADRF). First, we use Transformer-based encoding and decoding modules to describe the scene and predict future trajectories of surrounding HDVs through interactive modeling. After that, we model the driving intentions of surrounding drivers and predict the trajectories of HDVs based on the representation of vehicle interactions to construct the SADRF. Finally, trajectories can be planned online in low-risk areas of the SADRF to achieve safe and suitable autonomous driving.

S$^{\text{4}}$TP overcomes the limitations of short-term dynamic risks and the oversight of social interactions by considering the social interactions of other drivers and future traffic conditions, thereby enhancing the precision of risk assessment for AVs.
Furthermore, by employing SADRF-based trajectory planning, S$^{\text{4}}$TP emulates human driving patterns more authentically, enhancing the ride experience of autonomous driving.
The main contributions of this work are listed as follows:

\begin{enumerate}
    \item The S$^{\text{4}}$TP framework is proposed to balance the social suitability, safety, efficiency, and rationality in trajectory planning for AVs within complex, socially interactive traffic scenarios. The method employs SADRF-based condition constraints to generate optimized vehicle trajectories. The optimization algorithm is used to align with vehicle dynamics, such as acceleration and deceleration rates, turning radius, and vehicle stability, to ensure reasonable human-like planning control and maintain safety standards.

    \item We propose an effective SADRF algorithm that improves the safety and comfort of AVs in socially interactive traffic scenarios. SADRF considers the social dynamics of HDVs, predicts their possible trajectories, and generates a safer trajectory for the AV by assessing the surrounding risks. 
    Compared to traditional static DRF, SADRF more accurately assesses the risks of ego-circulation in a mixed traffic environment, particularly for AV itself.
   
    \item S$^{\text{4}}$TP have been validated in four challenging simulated driving scenarios, demonstrating superior performance in safety, efficiency, and social suitability compared to benchmark methods.
   
\end{enumerate}

The rest of the article is organized into the following sections:
Section \ref{sec:related_work} discusses the related literature works and emphasizes our contributions.
Section \ref{sec:method} describes the problem statement and the details of our proposed S$^{\text{4}}$TP method.
Section \ref{sec:expe} presents the experimental setup and results.
Finally, Section \ref{sec:conc} concludes the article.

\section{Related Works}\label{sec:related_work}

\subsection{Modeling of Driving Risk Field} 

The DRF is a driver model grounded in theories of traffic psychology and cognitive science, enabling it to accurately capture the subjective perceived risk of a driver in various driving scenarios \cite{15}. Establishing a DRF model enables autonomous driving systems to simulate driver behavior and make decisions more effectively, thus enhancing the safety and reliability of such systems.
The first research dates back to 1936 when Gibson\textit{ et al.}\cite{gibson1938theoretical} proposed that drivers perceive the qualitative concept of ``a field of safe travel" which consists of the possible paths that the car can take unimpeded. 
\textcolor{black}{This theory aims to describe the range within which drivers perceive their driving to be safe,  significantly contributing to improving safety and reliability in autonomous driving.} 
This pioneering theory laid the foundation for the ``motivational driver model". Subsequently, \textcolor{black}{Wilde \textit{et al.} \cite{wilde1982theory} proposed the risk steady-state theory, and Fuller \textit{et al.}\cite{18} introduced the task difficulty steady-state theory to elucidate driver behavior characteristics. These theories provide a solid foundation and invaluable perspectives that drive the progress of DRF modeling, fostering a more thorough and coherent understanding of driver behaviors and decision-making intricacies.}

In recent years, some studies have leaned towards delving deeper into modeling DRF.
For example, the concept of DRF was first proposed by
Kolekar\textit{ et al.}\cite{19}, who used a two-dimensional field that represents the belief of the driver. The DRF estimates the risk perceived by drivers, serving as a valuable tool in improving the safety and performance of autonomous driving systems. \nathaniel{However, its inability to account for real-time traffic scenarios and dynamically changing factors could lead to potentially inaccurate risk assessments.}
Moreover, in efforts to plan trajectories that minimize risk in traffic scenarios, Nyberg \textit{et al.} \cite{nyberg2021risk} and Zheng \textit{et al.} \cite{zheng2020bezier} integrated the driving risk by measuring the severity of safety constraints violations and modeled target uncertainty using potential field functions, respectively. Du \textit{et al.} \cite{du2022learning} proposed an innovative data augmentation that leverages DRF alongside original expert demonstrations to generate critical scenarios. This approach enables the model to more effectively learn critical driving behaviors, thereby enhancing driving policies and, as a result, improving the safety of AVS. However, a significant limitation in much of the current research on static DRF is the neglect of social interactions among drivers. This oversight hinders the dynamic adjustment of risk fields in scenarios involving social interactions, leading to a reduction in the accuracy of risk assessments for AVs.

To address the limitations of traditional DRF models, particularly their inadequacy in dynamically adjusting to real-time traffic scenarios and social interactions, we introduce the SADRF. This innovative approach is constructed to assess the expected levels of surrounding risk for AVs when interacting with HDVs, each characterized by unique social traits. Through the generation of two-dimensional heat maps centered on AV, SADRF accurately evaluates the expected task distribution such as turning, braking, and acceleration, depending on the traffic situation. By factoring in social interactions, it provides a nuanced understanding of the driving risk field across various driving scenarios, ensuring both safety and comfort in trajectory planning for autonomous driving.
\nathaniel{
\subsection{Trajectory Prediction}

Accurate trajectory prediction plays a critical role in autonomous driving \cite{ren2021safety}. Previous research primarily relied on historical trajectories and dynamic features of vehicles for prediction\cite{mozaffari2020deep}. These methods used machine learning or deep learning techniques to learn vehicle behavior patterns and forecast future trajectories by analyzing historical data. Early methods, for instance, used CNN-based \cite{xie2020motion} and RNN-based \cite{fu2016using} approaches to extract spatial features (using a bird's eye view) and time-series information (from historical trajectories), respectively. 
However, these methods often overlooked the interactions between vehicles and other traffic participants, failing to fully consider the influence of the entire traffic environment. Recent research has shifted focus to include social interactions among vehicles, incorporating them into trajectory prediction models. Graph neural networks \cite{li2019grip} have been utilized to capture interaction information through vectorization, and some methods\cite{23, liu2022interaction} have introduced game-theory processes to establish interaction models between vehicles.
However, these models are often considered ``black boxes" because of their complex structure and lack of interpretability. The attention-based mechanism approach \cite{messaoud2020attention} has recently gained significant attention among researchers. This strategy allows the model to quickly focus on essential factors by assigning higher weights or greater attention to them. One particular attention mechanism is the Transformer method \cite{zhang2022trajectory}, initially developed for natural language processing tasks but successfully adapted to the field of autonomous driving. In this article, we propose a data-driven, social-aware trajectory prediction method. The method uses the historical trajectories of the agents in the scene, the planned trajectory of the AV, and the map polylines as input. We employ a Transformer network to accurately model the social interactions among the vehicles, enabling the decoder can generate multimodal prediction trajectories that are safe and interpretable. 

\begin{figure*}[t]
      \centering
      \includegraphics[width=1\textwidth]{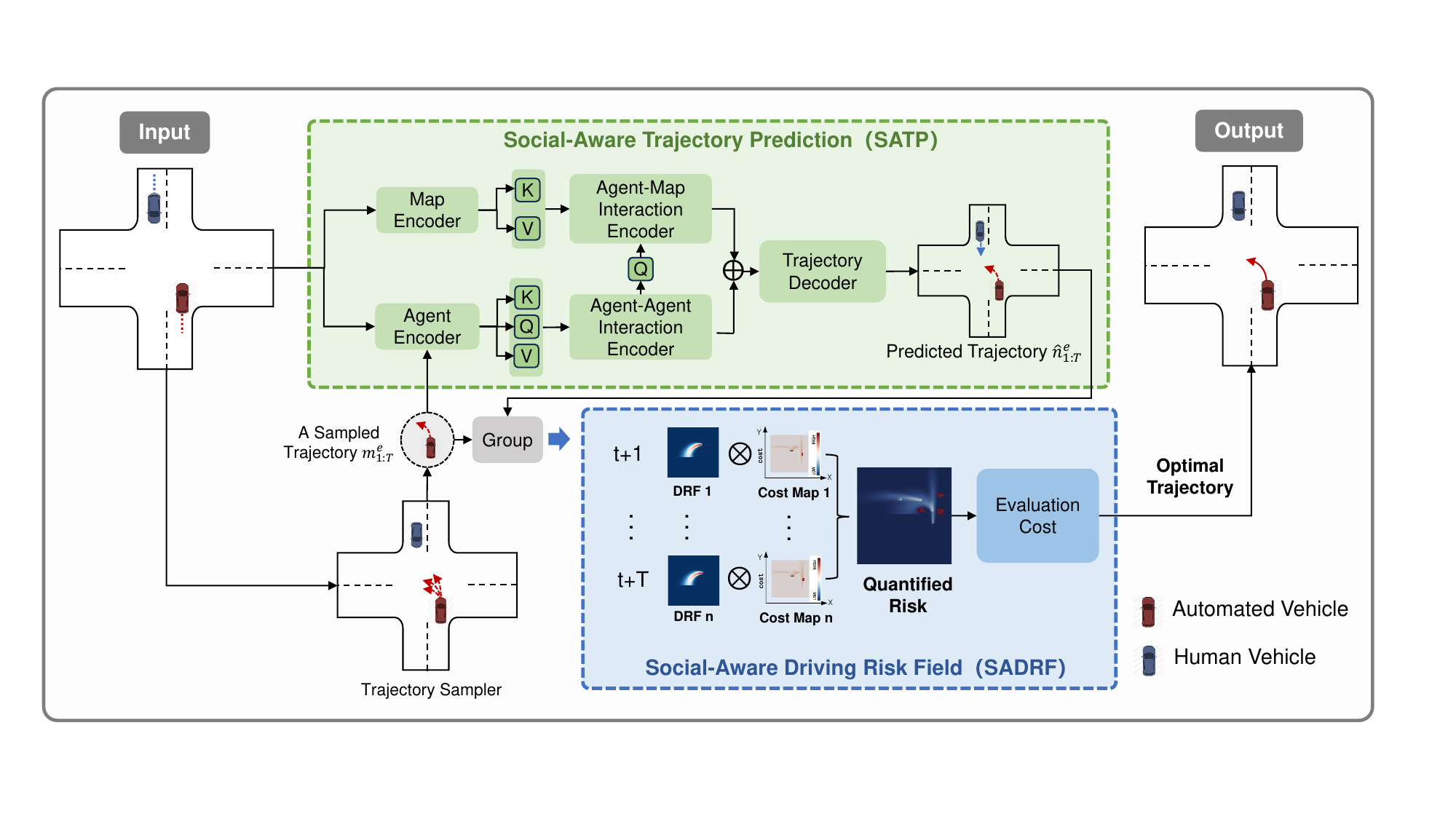}
      \caption{The overall architecture of the proposed S$^{\text{4}}$TP.} 
      \label{figurelabel}
\end{figure*}

\subsection{Trajectory Planning}

The trajectory planning module, as a core component of the autonomous driving system, receives environmental information from the perception module and generates planned trajectories for the downstream control module. In recent years, trajectory planning has been a subject of extensive research and exploration to cater to the demands of autonomous driving systems\cite{9712376}. Primary methods of trajectory planning include trajectory optimization, graph search, random sampling, and learning-based approaches\cite{huang2023differentiable}. Among these, learning-based and optimization-based methods are the most prevalent.

Learning-based methods utilize neural networks as driving policies to generate actions or trajectories directly from sensor inputs or perception results. Examples include imitation learning \cite{r1,torabi2018behavioral,ho2016generative} and deep reinforcement learning (DRL) \cite{2,schulman2017proximal,haarnoja2018soft}. Imitation learning involves observing expert behavior and learning their strategies to replicate similar behavior in the agent. It maps perception inputs to action spaces using neural networks to achieve motion planning. DRL refines the agent’s strategy through interactions with the environment, optimizing it by trial and error, guided by reward signals, to directly produce actions or trajectories. However, despite their simplicity and efficiency, learning-based methods using neural networks often face challenges in interpretability, generalization, stability, and safety\cite{kiran2021deep}.

To address these issues, we have chosen optimization-based methods for trajectory planning. Optimization-based methods offer better controllability and stability and can consider system constraints and limitations\cite{gonzalez2015review}. In our work, we implement real-time online trajectory planning, which involves executing the initial steps of the trajectory using a low-level motion controller and then re-planning. By scoring and evaluating the planned trajectories, we select the one with the highest safety score for execution by the downstream controller. Our approach differs from traditional optimization methods by integrating the predicted trajectories of surrounding vehicles, thereby enhancing the robustness of the system.}

\section{Methodology}\label{sec:method}

\subsection{Problem Statement}

We formulate trajectory planning as an optimization problem considering the risk of conflicts between AVs and surrounding HDVs and calculate an optimal trajectory based on safety, efficiency, and comfort indicators.

Assume that at a certain time $t$, the state of the AV is expressed as $m_t^e$, and the states of HDVs around are expressed as $n_t^e$.
$f$ represents the trajectory prediction model specifically designed to forecast the future state of HDVs.
Our model selects the optimal trajectory $\tau^\ast$ that minimizes the expected cost:   
\begin{equation} \label{1}
\tau^\ast=\underset{\tau \in \mathcal{O}}{\operatorname{argmin}} \ C(m_{1:T}^e,\ {\hat{n}}_{1:T}^e)
\end{equation}
\begin{equation} \label{2}
{\hat{n}}_{1:T}=f(n_{-T_h:0}^e,\Gamma,\ m_{1:T}^e)    
\end{equation}

In the Eq. (\ref{1}) and Eq. (\ref{2}), $m_{1:T}^e$ is a sequence of the AV's future states, which represents the optimal trajectory that we aim to optimize. 
\nathaniel{$T$ refers to the planned trajectory range, }$T_h$ is the historical trajectory range and $\Gamma$ represents the map information. $\mathcal{O}$ represents the collection of viable trajectories that are generated by considering the initial state $m_t^e$ of the AV.
$C$ is the cost function. The prediction result can have two forms: one is a distribution of possible future states per HDV; the other is a prediction of the joint trajectory for multiple HDVs in the future.

\subsection{Overall Architecture of S$^{\text{4}}$TP}

S$^{\text{4}}$TP is inspired by \cite{19,huang2023learning} and adopts the network architecture shown in Fig. \ref{figurelabel} to perform decision-making and planning tasks.  
Our proposed framework S$^{\text{4}}$TP contains two parts, namely social-aware trajectory prediction (SATP) and social-aware driving risk field (SADRF). SATP is designed primarily on the basis of the constructed scene representation and multiple trajectories planned by the AV, which allows for the prediction of trajectories of surrounding agents. 
Specifically, we adopt Transformer-based encoding modules to model the agent-map and agent-agent interactions in the scene. \nathaniel{Combining the sampled trajectories of the AV, we utilize a  Gated Recurrent Unit (GRU) to autoregressively decode the AV's future states, thereby generating predicted trajectories for the HDVs in the scene. The SADRF module then utilizes the predicted trajectories from SATP and the sampled AV trajectories to model the dynamic driving risks around the scene. Subsequently, real-time online trajectory planning is carried out within the risk field using an optimization-based approach.}

\nathaniel{S$^{\text{4}}$TP method is trained online and in real-time using the SATP module, ensuring that the agent can autonomously explore and elicit responses from surrounding agents. At the same time, the model stores the interactions between vehicles in a replay buffer to effectively train the prediction model. AV planning is carried out using the SADRF module.}
Based on the SADRF, we plan trajectories online in low-risk areas to generate safe, comfortable, and socially appropriate driving trajectories that comply with human driving preferences, which are used for execution.

\begin{figure*}[t]
      \centering
      \includegraphics[width=0.75\textwidth]{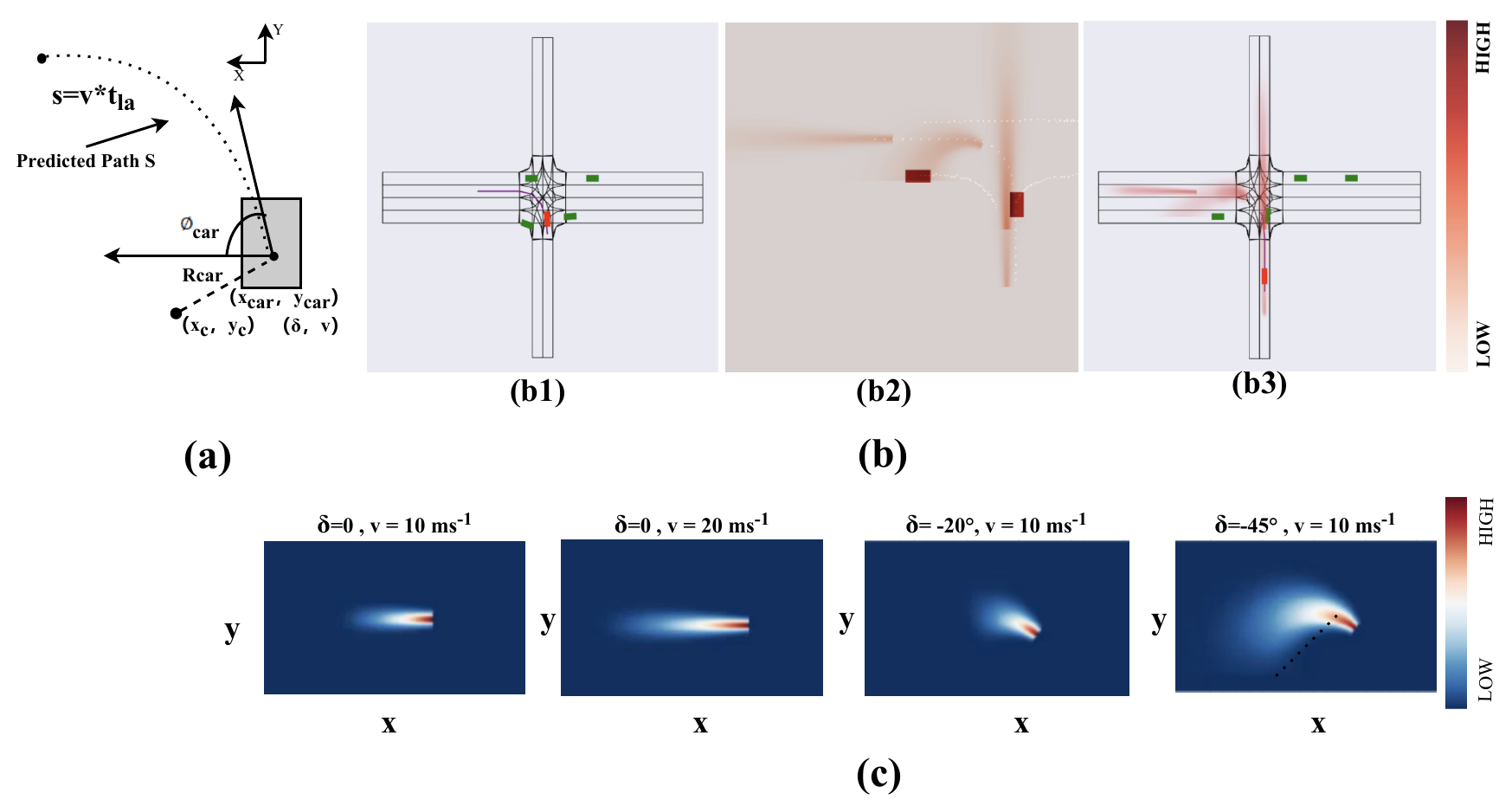}
      \caption{\textbf{Illustration of modeling the SADRF.} To implement our proposed SADRF modeling, we begin by calculating the predicted trajectories of HDVs based on the vehicle dynamics model, \nathaniel{as depicted in (a), where we use the state inputs as the position($x_{\text{car}},y_{\text{car}}$), heading $\phi_{\text{car}}$ and the steering angle $\delta$ of HDVs. \nathaniel{The predicted path (arc length $s$) is calculated by $s =v*t_{\text{la}} $.} 
      (b) Showcases the dynamic generation of DRF, specifically during the planning of an unprotected left-turn path. (b1), (b2), and (b3) illustrate path planning process, visualization of SADRF alone, and visualization of SADRF incorporated into the scene, respectively.} The individual SADRF graph on the right side represents the risk values within the scene using a color-coded bar chart ranging from light to dark red. In this visualization, the highest risk value is assigned to HDVs. The risk value is dynamically computed based on the currently planned trajectory, adapting to changes at subsequent moments during dynamic planning. 
      (c) Demonstrate the shape of the SADRF with respect to velocity $v$ and steering angle $\delta$. As can be seen, the SADRF expands as the velocity increases and the steering angle increases, thus improving the security for subsequent trajectory planning.}
\label{overall}.
\end{figure*}

\subsection{Social-Aware Trajectory Prediction}\label{sec:prediction}
 
 \nathaniel{The SATP consists of a scene encoder and a trajectory prediction decoder, which collaborate to generate accurate predictions for HDV trajectories.
 First, the scene encoder takes the historical trajectories of surrounding agents and local map polylines as inputs. Then, we use the multi-head attention mechanism in the Transformer network to accurately encode and extract features for the intersection of agent-agent and agent-map.}
 Finally, the trajectory prediction decoder fully considers the social interactions of surrounding HDVs and uses a self-recursive GRU to decode the future states of HDVs, thereby achieving interaction prediction of surrounding vehicles. The scene encoder and trajectory prediction decoder are the two main parts of a prediction model and are detailed as follows.

\textbf{Scene encoder:} 

We adopt an encoder structure similar to \cite{huang2022multi,huang2023differentiable}, and the scene encoder is based on Transformer networks. For each HDV $i$ around the AV, the encoder input data includes the historical state $n_{-T_h:0}^i$ of the HDV and the polylines of the local map $\Gamma  ^{i} $ (that is, a list of route points in nearby locations). Additionally, we explicitly consider the interaction relationship between the AV and surrounding HDVs, adding the historical state information $n_{-T_h:0}^e$ of the AV to the input data to construct a complete interaction graph.
To maintain the fixed shape of the tensor, we handle missing time steps in the historical trajectory and missing points in the polyline of the map by filling them with zero values.
In the experiments, our model selects a certain number (set to 5 in the experiments) of HDVs close to the AV and performs zero padding when the number is insufficient. The original input map polyline is encoded through a multilayer perceptron (MLP). In order to capture the temporal relationship in the historical trajectory of other HDVs, we utilize the Transformer-based self-attention layer for encoding. \nathaniel{As depicted in Fig. \ref{figurelabel}, we employ a hierarchical Transformer module to encode the interactions relationships between the agent-agent and the agent-map, thus extracting rich contextual information. In particular, we represent all agents as nodes in the constructed interaction graph and use a two-layer self-attention Transformer encoder as the interaction encoder to handle the interaction relationships between agent-agent in the graph\cite{tian2022federated}, where query, key and value ($Q$, $K$, and $V$) are encoded historical trajectory features of agents. To capture the agent-map interaction, we employ two cross-attention Transformers for encoding, separately modeling the agents on the lane vectors in the map and the agents on the sidewalk vectors in the map. We utilize the interaction features of the agents as query $Q$ and use a single map vector (encoding the sequence of path points) as key and value ($K$ and $V$). Finally, we concatenate the historical features, interaction features, and map attention features of the agents in the scene, and pass them through a trajectory prediction decoder to generate predicted future trajectories.}


\textbf{Trajectory prediction decoder:} 
Our decoder combines the sampled trajectory of the AV to predict the future trajectories of surrounding HDVs. Compared to other recurrent neural networks, GRU has fewer parameters and does not require an explicit memory unit, so it performs better when processing long sequences \cite{wei2023stgsa}. We use GRU in the model to decode the state sequence of HDV over a period of time in the future through autoregressive means. 
To explicitly model the impact of AV on the future actions of surrounding HDVs, we input the motion state $m_t^e$ of AV at time $t$ into the GRU to simulate the impact of AV on the future state of HDVs and make interaction-aware prediction. 
Specifically, at each moment in the future, the GRU accepts the hidden state of the previous moment and the joint input representation of the last predicted state and the AV state at that moment. 
By updating the hidden state to decode the change in the HDV state (including the coordinates $x$ and $y$, and the heading angle $\phi$), which adds up to the last state at the previous moment, the predicted state is obtained at the current time. 
Additionally, we consider the influence of the AV on each surrounding HDV that is not in the conflict zone. Furthermore, we also introduce the interactive gating network $\mathbb{G}$ to model the pairwise interactions between the AV and the target HDV in the intersection scene. \nathaniel{$\mathbb{G}$ is an MLP that takes concatenated features from the historical encoder of the AV and the target HDV as input. The output scores, obtained through a sigmoid function, represent the interaction strength between AV and HDV. This interaction strength is applied to the input of the AV for future states and passes through the GRU unit.} The trajectory prediction decoder outputs its future multimodal trajectories for each HDV that participates in the same scenario. \nathaniel{For the AV, we decode its future control actions (i.e., acceleration $a$ and steering angle $\delta$) from the feature vectors using an MLP. We then utilize a kinematic model to convert the sequence of actions into trajectories.}

\subsection{Social-Aware Driving Risk Field}

To effectively evaluate the surrounding driving risk in complex interaction scenarios, 
we construct an SADRF model that uses the predicted trajectories of the surrounding HDVs and the sampling trajectories of the AV. 
Our model considers the social interactive relationships between AV and HDVs in complex traffic scenarios and evaluates future risks, providing valuable support for AV planning and decision-making.

The proposed SADRF model is constructed on the basis of DRF theory \cite{19}. 
DRF model builds a 2D Gaussian distribution along the predicted path to represent the subjective perception of HDVs in the surrounding environment\cite{du2022learning}.
 \nathaniel{
The heights $a$ and width $\sigma$ of the Gaussian distribution are a function of the arc length $s$. The length of the arc refers to the length of the trajectory followed by the AV.
The predicted path and the corresponding Gaussian in each cross section along the predicted path are defined by the equation:}

\begin{equation}\label{eq:drf}
G(x, y)=a(s) \exp \left(-\frac{\sqrt{\left(x-x_c\right)^2+\left(y-y_c\right)^2}-R_{\text{car}}}{2 \sigma^2}\right) 
 \end{equation}
\begin{equation}\label{car}
R_{\text{car}}=\frac{L}{\tan \delta}
\end{equation}
In Eq. (\ref{eq:drf}) and Eq. (\ref{car}), $R_\text{car}$ is the predicted trajectory radius calculated according to the vehicle wheelbase $L$ and the steering angle $\delta$.\nathaniel{ In this article, we calculate the predicted trajectory of the vehicle using a kinematic model. As shown in Fig. \ref{overall}(a), the rotation center$(x_c,y_c)$ of the vehicle is determined using position ($x_{\text{car}},y_{\text{car}}$), heading ($\phi_{\text{car}}$). This center is then used to calculate the arc length $s$ measured along the predicted path.} 
 The Gaussian distribution $G(x,y)$ assigns probabilities to possible locations of the AV in the next step, based on its values at each $(x,y)$ coordinate. Height $a$ is modeled as a parabola:
\begin{equation}\label{a}
a(s)=p\left(s-vt_{\text{la}}\right)^2 
\end{equation}
In Eq. (\ref{a}), $t_{\text{la}}$ is a fixed look-ahead time, and the look-ahead distance is assumed to increase linearly with speed $v$. The parameter $p$ determines the slope or ``steepness" of the Gaussian parabolic curve.
\nathaniel{The width $\sigma$ is modeled as a linear function of arc length $s$, which refers to the length of trajectory that AVs follow:
}
\begin{equation}\label{width} 
\sigma_i=\left(m+k_i|\delta|\right) s+c
\end{equation}
\begin{equation}\label{i} 
i=
\left\{  
             \begin{array}{lr}  
             1 ,
               & \text{inner} \ \nathaniel{\sigma}\\  
             2,
             & \text{outer} \ \nathaniel{\sigma} \\  
             \end{array}  
\right. 
\end{equation}

In Eq. (\ref{width}) and Eq. (\ref{i}), $m$ defines the slope of widening of the DRF when $\delta=0$ (driving straight). $k_{1}$ and $k_{2}$ exhibit a proportional relationship to the width of the DRF, where they increase (or decrease) as the absolute steering angle $\delta$ increases. 
\nathaniel{The parameter $c$ is defined as the width of the DRF at the location of the vehicle and is directly related to the car width. In this article, $c$ is equal to car-width/ 4, which corresponds to ±2 $\sigma$ of a Gaussian distribution, accounting for 95\% of the distribution's probability density \cite{19}.
}The parameters $k_{1}$ and $k_{2}$ are utilized to define the inner and outer edges of the DRF, respectively, enabling the representation of an asymmetric DRF. The DRF expands when both $k_{1}$ and $k_{2}$ are positive and contracts when both are negative.
In summary, the DRF is characterized by parameters such as $p$, $t_{l a}$, $m$, $c$, $k_{1}$, $k_{2}$, and is dependent only on the driver's state, without taking into account the surrounding environment. \nathaniel{In the same traffic scenario, drivers with larger parameter values of the DRF tend to perceive a higher level of risk compared to those with smaller values. Besides, the objective cost map of the environment is unaffected by the driver's subjective perspective, ensuring consistency across all individuals. Consequently, the parameter settings are kept consistent for all vehicles in this work.

}

\begin{figure*}[t!]    
    \centering
    \begin{subfigure}[b]{0.34\textwidth}
           \centering
           \includegraphics[width=\textwidth]{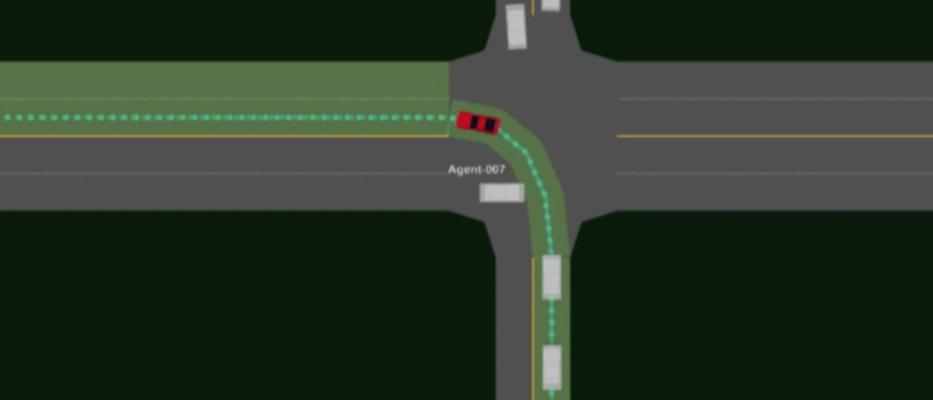}
            \caption{Left-turn scenario}
            \label{Fig:a}
    \end{subfigure}
    \begin{subfigure}[b]{0.34\textwidth}
            \centering
            \includegraphics[width=\textwidth]{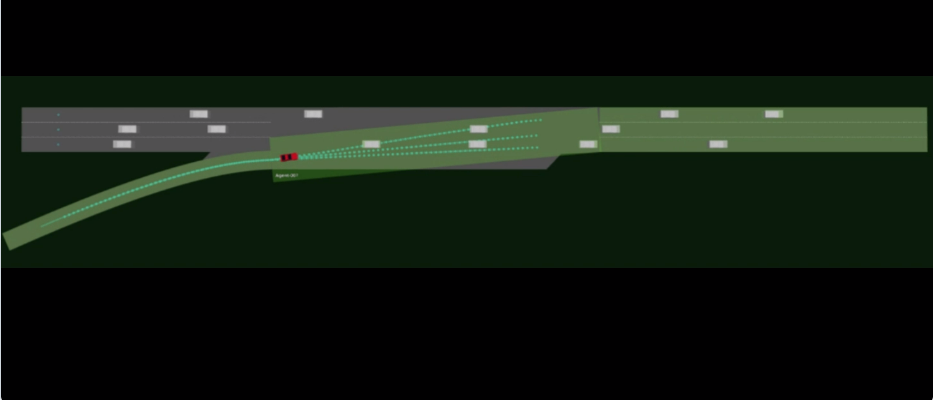}
            \caption{Merge scenario}
            \label{Fig:b}
    \end{subfigure}
    \begin{subfigure}[b]{0.34\textwidth}
            \centering
            \includegraphics[width=\textwidth]{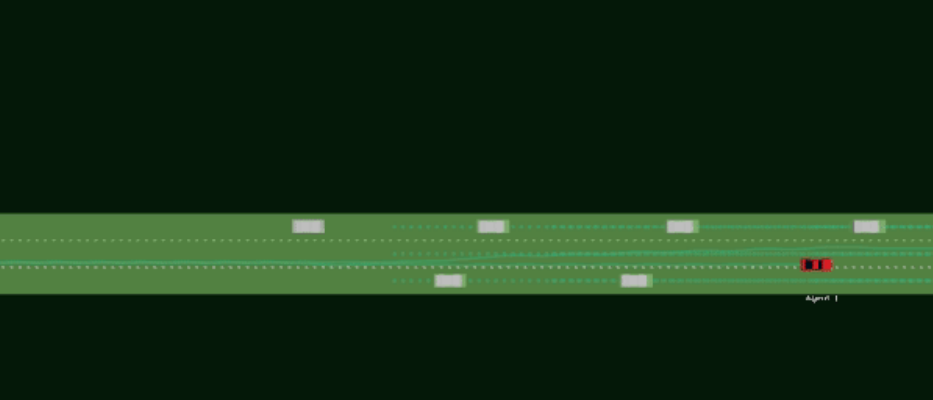}
            \caption{Cruise scenario}
            \label{Fig:c}
    \end{subfigure}
    \begin{subfigure}[b]{0.34\textwidth}
           \centering
           \includegraphics[width=\textwidth]{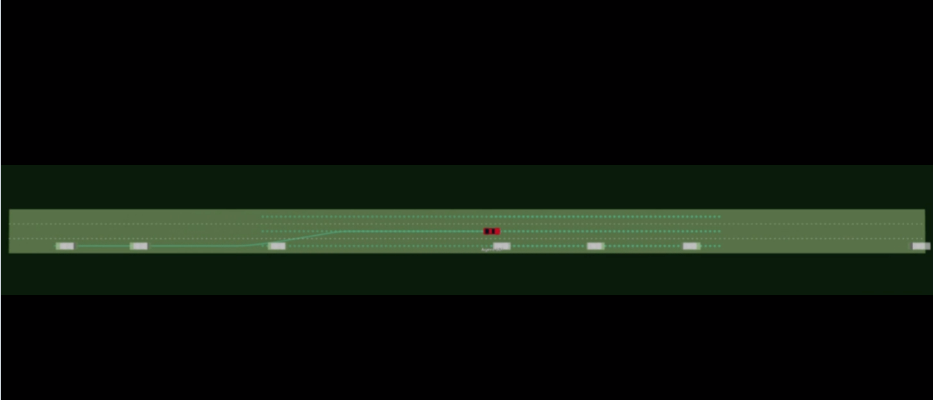}
            \caption{Overtake scenario}
            \label{Fig:d}
    \end{subfigure}
    \caption{Four designed driving scenarios in the SMARTS environment.\centering}
    \label{fig:test_scenario}
\end{figure*}

To calculate future actions for the DRF model, SADRF adopts an approach similar to \cite{19}, where the perceived risk is calculated based on the current velocity $v$ and the steering angle $\delta$ of the AV.
Then, an optimization process is carried out based on the risk threshold theory to determine the future velocity and steering angle that keep the perceived risk below the assigned threshold.

Our SADRF can dynamically adjust at the time step $t$ of the AV and matches the dynamic parameters of different risk field models according to different speeds $v$ based on the existing static DRF\cite{19}. This enables SADRF to have the ability to dynamically assess HDV risk levels. The construction of SADRF includes the following steps:

\nathaniel{First, we predict the HDV trajectories at the future time step $t$ and construct a static DRF based on the AV pose. When constructing the static DRF, we consider factors such as the speed $v$ and direction $\phi$ of other vehicles to improve the accuracy of the predictions. This process depends on Eq. \ref{eq:drf} to be completed.}

Second, we add the social interactions of surrounding HDVs through the interactive prediction model described in Subsection \ref{sec:prediction} and obtain the multimodal trajectories of HDVs, thus obtaining the pose prediction results of other vehicles. As illustrated in \cite{19}, we know that the perceived risk of the driver is the convolution product of the subjective DRF map $\nathaniel{G(x, y)}$ of the HDVs and the objective cost map $\nathaniel{M(x, y)}$ of the traffic environment.  
\nathaniel{To construct the cost map, the AV considers other vehicles as static obstacles at time $t$. For each obstacle, based on its dynamic risk characteristics (such as distance to the AV, velocity difference, etc.), it assigns an appropriate risk level. The cost map denoted $M(x, y)$, is built by combining the risk levels of all obstacles with the geometry of the road map.
It is important to note that the information from the cost map can be used for both path planning and path evaluation purposes. It enables the system to select paths with lower risk levels, thus aiding in safer navigation and decision-making.}
Therefore, we consider the predicted results of the HDV poses as static obstacles at time $t$, and use the road map to construct a static cost map $M_t(x_t,y_t)$ at time $t$. \nathaniel{For each sampled AV trajectory, we obtain a series of risk field maps $G(x_t,y_t)$ and cost maps $M_t(x_t,y_t)$ at different times in the future, and based on this, we can generate a Social-Aware Driving Risk Field $G_t(x_t,y_t)$
based on social interactions between AV and HDV.}
Thus, by obtaining a series of risk field maps and cost maps, the AV can better understand the safety and driving conditions of the surrounding environment and make more accurate and efficient decisions.

\nathaniel{Finally, for any time step $t$, we can obtain the total risk threshold $R_{\text{risk}}(t)$,
which can be calculated by multiplying the cost map and the SADRF. 
The expression is as follows:
\begin{equation}\label{4}
R_{\text{risk}}(t)=\sum_{t=1}^{T}{G_t(x_t,y_t)\ast M_t(x_t,y_t)}
\end{equation}
In the Eq. (\ref{4}), $T$ represents the total number of time steps. 
It is important to note that both the risk field and the cost map dynamically change with the speed $v$ and the steering angle $\delta$ of the AV. Therefore, the total risk value calculated at each time step will be different. The AV needs to continuously update the risk field and the cost map based on new data information and calculate the total risk value to make final decisions in real time.}

We visualized the correspondence of our proposed SADRF model with speed and steering angle through Gaussian plots, which is shown in Fig. \ref{overall}(c). \nathaniel{As the speed of the AV increases, the range of the SADRF expands accordingly. Similarly, with an increase in the steering angle, the SADRF also expands to predict potential risks}


\subsection{SADRF-Based Trajectory Planning}


By constructing the SADRF, as demonstrated in Fig. \ref{overall}(b), we map the surrounding risk factors of the AV ( such as other vehicles, pedestrians, etc.) into a two-dimensional heat map representation of the risk field and use the sum of the total risk within a certain future time period as a standard to evaluate the safety of the AV trajectory. \nathaniel{Subsequently, the AV performs online trajectory planning within low-risk areas. In the planned trajectory, a scoring calculation is conducted using preferred trajectories. The trajectory with the highest safety score is selected for execution by the downstream controller, resulting in safer automated driving decision-making and planning.
Compared to the traditional static DRF method, S$^{\text{4}}$TP models the risk of the AV trajectory in a more granular way. This intuitively represents the distribution of risk in the environment, achieving a safer and more robust AV trajectory planning. Furthermore, our method
avoids generating overly conservative trajectories, ensuring that AVs can swiftly pass through intricate social interaction scenarios.}

\section{Experiments}\label{sec:expe}

\subsection{Experimental Setup}


As depicted in Fig. \ref{fig:test_scenario}, we applied four typical autonomous driving scenarios within the SMARTS simulator \cite{zhou2020smarts} to assess the effectiveness of the proposed method as follows.
(1) \textbf{Unprotected left-turn}: \textcolor{black}{In the absence of signals or protection, the AVs encounter traffic from different directions and must coordinate with other HDVs to navigate the intersection safely and efficiently.}
(2) \textbf{Intersection merging}: The AV needs to seamlessly merge into traffic flow at an intersection where two or more roads converge and navigate safely through dense traffic on the main road.
(3) \textbf{Road cruising}: \textcolor{black}{The AV navigates along a predetermined road route while abiding by traffic regulations and safety laws}. 
(4) \textbf{Lane overtaking}: The AV performs a maneuver of passing a vehicle ahead while driving.
To ensure fairness in the comparative experiments, we conducted 500 episode tests for each scenario. 
\nathaniel{Episodic data is stored in a replay buffer, and batch samples are subsequently drawn from this buffer to train the motion predictor. }

\begin{table*}[t]
  \centering
  \caption{Performance comparison of different trajectory planning methods in various scenarios}
  \label{tab:transposed}
  \renewcommand{\arraystretch}{1.1}
  \resizebox{2\columnwidth}{!}{    
  \begin{tabular}{ll|cccc|ccc}
    \toprule
    \textbf{Scenario} & \textbf{Metric} & \textbf{Fanta} & \textbf{SMARTS} & \textbf{Discrete-SAC} & \textbf{Safety-Balanced} & \textbf{Predictive-Decision} & \textbf{Static DRF} & \textbf{S$^{\text{4}}$TP (Ours)} \\
    \midrule
    \multirow{3}{*}{Left Turn} 
    & Success $\uparrow$ & 94\% & 85\% & 88\%  &89\% & 89\% & 80\% & \textbf{100\%} \\
    & Time $\downarrow$ & 187.68 & 235.99 & 251.43  &200.51 & 148.67 & 241.74 & \textbf{22.60} \\
    & Humanness $\downarrow$ & 2148.53 & 1739.49 & \textbf{35.05}  &1132.39 & 920.48 & 1047.13 & 515.69 \\
    \midrule
    \multirow{3}{*}{Merge Lane} 
    & Success $\uparrow$ & \textbf{100\%} & 4\% & 98\%  &90\% & 91\% & 79\% & \textbf{100\%} \\
    & Time $\downarrow$ & 168.92 & 831.56 & 161.20  &178.16 & 130.77 & 245.80 & \textbf{46.26} \\
    & Humanness $\downarrow$ & 3195.57 & 967.74 & 16.87  &78.08 & 360.09 & 1191.81 & \textbf{4.77} \\
    \midrule
    \multirow{3}{*}{Cruise} 
    & Success $\uparrow$ & 99\% & 80\% & \textbf{100\%}  &94\% & 99\% & \textbf{100\%} & \textbf{100\%} \\
    & Time $\downarrow$ & 161.62 & 363.81 & 150.04  &154.03 & 55.69 & 60.16 & \textbf{46.07} \\
    & Humanness $\downarrow$ & 872.11 & 939.05 & 31.29  &34.65 & 14.36 & 25.32 & \textbf{0.00} \\
    \midrule
    \multirow{3}{*}{Overtake} 
    & Success $\uparrow$ & \textbf{100\%} & 66\% & 97\%  &98\% & \textbf{100\%} & \textbf{100\%} & \textbf{100\%} \\
    & Time $\downarrow$ & 168.92 & 465.91 & 215.44  &118.66 & 60.16 & \textbf{57.82} & \textbf{57.82} \\
    & Humanness $\downarrow$ & 3195.57 & 940.43 & 26.19  &28.21 & 25.32 & \textbf{16.54} & \textbf{16.54} \\
    \midrule
    \multirow{1}{*}{All} 
    & Overall Success $\uparrow$ & 98.25\% & 58.75\% & 95.75\%  &92.75\% & 94.75\% & 89.75\% & \textbf{100\%} \\
    \bottomrule
  \end{tabular}
  \label{tab:compare}
}
\end{table*}

\subsection{Baselines}
We compared the proposed S$^{\text{4}}$TP with the following strong baselines:
\begin{enumerate}    
    \item Fanta\cite{mahierarchical}: This is the first-place solution in the NeurIPS 2022 autonomous driving competition. It proposes a hierarchical architecture, where the upper-level module (meta-controller) decides whether the AV should move or remain stationary, trained using supervised and DRL. The lower-level module includes three policies used to determine the AV's speed, heading, and whether to switch to a different lane.    
    \item SMARTS \cite{zhou2020smarts}: It is a solution that considers multi-HDV DRL with social interaction for trajectory planning, which can serve as a benchmark method for DRL and multi-HDV autonomous driving research. This approach focuses on diverse interaction experiences.
    \item Discrete-SAC  \cite{haarnoja2018soft}: This method trains a multitask RL AV and a scene classifier. For multitask RL training, the SoftModule network is used as the backbone architecture for the policy networks and Q-networks of the SAC algorithm. The MLP scene classifier receives observations as input and generates a one-hot encoding to determine which task module to be utilize.
    \item \textcolor{black}{Predictive-Decision \cite{huang2023learning}: This is the third-place solution in the NeurIPS 2022 autonomous driving competition. It proposes an interactive perception motion prediction model that uses a Transformer network to encode driving scenes and predicts other HDVs' future trajectories in combination with the AV's future planning.}
    \item  Static DRF \cite{19}: This is a method based on a static driving risk field to generate a predictive model of human driving behavior. This method models static and dynamic obstacles in the scene to construct a static driving risk field. Then, by combining vehicle position and velocity information with a static driving risk field, a risk measurement is generated, which can be converted into vehicle control operations, such as acceleration, deceleration, and steering for decision-making.
    \nathaniel{\item Safety-Balance \cite{wang2023safety}: This method utilizes a mixture-of-experts approach and combines it with a Transformer model to achieve scene-consistent multimodal trajectory prediction. It also facilitates safe and efficient trajectory planning by considering uncertain social circumstances and interactions with vehicles of different driving styles.}
\end{enumerate}

Predictive-Decision and static DRF are used in ablation experiments, where Predictive-Decision is the method without adding SADRF, and static DRF replaces the SADRF module with a static risk field.

\subsection{Implementation Details}
\textcolor{black}{In order to ensure comparability, we adopted the experimental settings introduced in \cite{huang2023learning}, which promote a fairer evaluation of results. Our framework uses a prediction and planning time horizon of 3 seconds($T = 30 $), a historical observation range for the HDV of 1 second($T_h = 10)$ with a time interval of 0.1 seconds. We select the five HDVs closest to the AV and predict their future trajectories. The polyline map input to the model includes 50 waypoints for each path taken by HDVs, with a separation of 1 meter between each waypoint.
Our model is implemented in PyTorch \cite{pytorch} and trained on a NVIDIA V100 GPU. We used a batch size of 64 and trained with Adam optimizer with an initial learning rate of 2e-4 and a decay factor of 0.9 every 4000 training steps. There are 1000 training episodes, each followed by 50 gradient steps. Our DRF model-related parameter settings are consistent with those in \cite{19}. For the hyperparameters, we set $p=0.0064$, $t_{la}=3.5$, $c=0.5$, $m=0.05$, $k_1=0.5$ and $k_2=1.38$ in Eq. (\ref{width}).
}
\subsection{Evaluation Metric}

\begin{figure*}[t]
      \centering
      \includegraphics[width=1\textwidth]{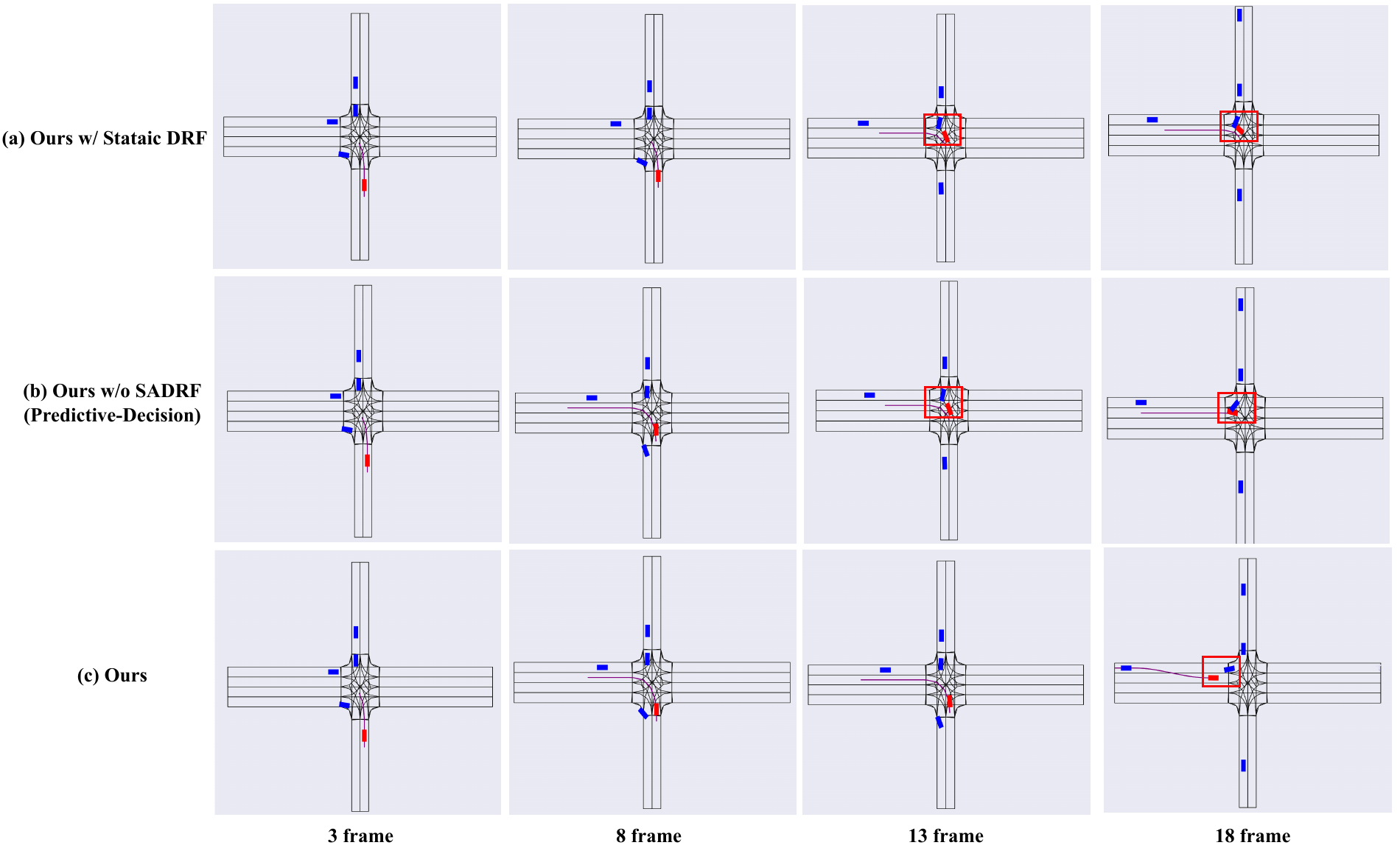}
      \caption{\nathaniel{Visual comparison of planning trajectories between our proposed S$^{\text{4}}$TP, the baseline Prediction-Decision, and the static DRF in unprotected left-turn intersection scenario. The red box and line are the AV and its planned trajectory; the blue boxes are the surrounding agents.
      (a) The proposed SADRF-based planner excels at accurately assessing surrounding risk levels in complex social interaction scenarios, thus ensuring the safety of the AV. In contrast, (b) the Prediction-Decision planner, which may disregard risk fields, fails to accurately assess the surrounding risk levels and consequently results in collisions. (c) The static DRF, lacking the ability to dynamically assess trajectory risk values, leads to overly conservative strategy generation in complex scenarios compared to SADRF.
      }}
      
      \label{vis1}
\end{figure*}

The compared trajectory planning methods will be evaluated on the following indicators:

\begin{enumerate}
    \item \textbf{Success}: Complete goal rate. This metric is calculated by dividing the total number of non-collisions by the total number of episodes, and a bigger value indicates a lower frequency of failure instances.
    \item \textbf{Time}: Number of steps taken and final distance to the goal. This metric is calculated by dividing the cumulative number of AV adjustments in each episode and the distance to the destination by the total number of episodes.
    \item \textbf{Humanness}: Similarity to human behavior. This metric is obtained by dividing the cumulative distance measurements by the obstacle, the cost of the jerk degree, and the cost of road offset by the total number of episodes in all vehicles. \nathaniel{The calculation formula is as follows.
    \begin{equation}
        H=C(d_\text{ob}+j_\text{an}+j_\text{in}+l_\text{cen})/n_\text{all}
    \end{equation}
    where $d_\text{ob}$ represents the distance between the AV and obstacles, $j_\text{an}$ represents the angular jerk, $j_\text{in}$ represents the linear jerk, $l_\text{cen}$ represents the distance of the AV from the center of the lane, $n_\text{all}$ represents the total number of vehicles in the scene, $C$ represents the cost function, and $H$ represents the humanness metric, which can be used to assess the degree of human-like behavior exhibited by the vehicle. A lower value indicates a behavior that is more human-like.}
\end{enumerate}

\subsection{Comparison Results}

We conduct a comparison between our proposed method and several benchmark methods. To evaluate the performance of S$^{\text{4}}$TP, we conduct performance tests by introducing an increase in the diversity of traffic flows within the scenario.  
Table \ref{tab:compare} displays the evaluation results of 500 different traffic flow patterns, which indicate that our method significantly outperforms RL or other methods in terms of both success rate and efficiency. 
\nathaniel{According to Table \ref{tab:compare}, in unprotected left-turn intersections, AV needs to coordinate with HDV coming from different directions to navigate the intersection safely. However, the static DRF method relies solely on historical data and geographical information for scene encoding, without considering the driver's real-time perception and specific social preferences. This limitation prevents the method from dynamically recording various complex situations and changes in the real driving environment, resulting in a lower success rate.
SMARTS and Discrete-SAC employ traditional DRL methods and have achieved some level of success. However, due to the inherent drawbacks of DRL methods, their models exhibit limited generalization ability in complex interactive scenarios.
The Safety-Balance method takes into account the driving styles of surrounding HDVs and uses a Transformer-based encoding-decoding network to model the interactions between vehicles. Although this method performs well in the scenario of an unprotected left turn, particularly focusing on interactions at crossroads, its performance is average when compared to the other three scenarios.
The Fanta method combines supervised learning and reinforcement learning for training and incorporates a traditional rule-based collision detection module for self-vehicle decision planning, resulting in good results.
In comparison, our approach improves upon the Predictive-Decision method by introducing SADRF during the self-vehicle planning phase. This allows us to dynamically adaptively model the risks in the scene and accurately assess the driving risks around the self-vehicle, leading to better interpretability and generalization. Therefore, our method achieves the best results when compared to other methods.
Taking advantage of rich SADRF information, the proposed S$^{\text{4}}$TP achieves a success rate of 100\% in all test scenarios, while the state-of-the-art (SOTA) algorithm Fanta \cite{mahierarchical} exhibits a success rate of 98. 25\% and the predictive decision exhibits a success rate of 89.75\% in all scenarios. Furthermore, for other intersection scenarios, such as merge lane, cruise, and overtake scenarios, our method also achieved the best experimental results in the evaluation metrics of ``time" and ``humanness". This further demonstrates that S$^{\text{4}}$TP can achieve safer and more comfortable decision planning for autonomous driving.}
In summary, our method achieves the best performance across all scenarios compared to other SOTA approaches.

\subsection{Ablation Study}
\nathaniel{Fig. \ref{vis1} illustrates the visualization results while comparing our SADRF method with Predictive-Decision\cite{huang2023learning} and static DRF \cite{19} in a specific unprotected left-turn intersection scenario. For clearer differentiation, the trajectory planning processes are showcased at the 3rd, 8th, 13th, and 18th frames. Besides, any instances of collisions are emphasized and distinctly marked. Among them, Predictive-Decision operates without adding SADRF, and static DRF substitutes SADRF with static risk field. 
As shown in Fig. \ref{vis1}(a), the strategy generated by static DRF \cite{19} is overly conservative, leading to a significantly slower decision making process when passing through the intersection. Moreover, it encounters substantial difficulty during restarts, ultimately leading to a collision with an oncoming HDV.
Simultaneously, Fig. \ref{vis1}(b) demonstrates the results from the 8th to 18th frames, it becomes evident that after completing trajectory planning, Predictive Decision\cite{huang2023learning} struggles to adjust its strategy in real-time when confronted with aggressive or previously unnoticed vehicles (vehicles not observed or obscured) suddenly emerging in the driving scene, resulting in a collision with a human-driven vehicle. 
In contrast, as demonstrated in Fig. \ref{vis1}(c), S$^{\text{4}}$TP allocates a pre-set margin for the vehicle prior to planning to anticipate unforeseen circumstances. Unlike conservative strategies, S$^{\text{4}}$TP dynamically adapts to the movement of other HDVs within the scene in real-time. This can ensure the avoidance of dangerous situations at a specific future moment and generate more human-like driving trajectories. }

It should be noted that our method incorporates a driving risk field model, which allows it to dynamically adjust the driving risk field considering real-time road conditions and vehicle status. This adaptive capacity enables better prediction and avoidance of potential risks and hazards, resulting in metrics that exhibit improved human likeness. The quantitative results further substantiate that this innovative feature improves the adaptability of our method in real-world road scenarios, providing a higher level of safety assurance for autonomous driving.

\section{Conclusion}\label{sec:conc}
 
\textcolor{black}{In this article, we propose a social-suitable and safety-sensitive trajectory planning method, named S$^{\text{4}}$TP. This method considers the risks associated with social interaction in complex mixed traffic environments, ensuring safe, efficient, and socially appropriate trajectory planning for AVs.}
The proposed method employs Transformer-based encoding and decoding modules for scene representation and predicts future trajectories of surrounding HDVs using interactive modeling.
Based on these predicted multistep trajectories, an SADRF model is constructed to produce autonomous driving trajectories aligned with human driving behavior. Our approach utilizes multi-task learning in an end-to-end framework facilitating both training and evaluation in simulated driving scenarios.  
The results demonstrate the superiority of our approach over benchmark methods. With the SADRF module, we achieve higher success rates compared to traditional static DRF models or trajectory planning methods that neglect social interaction.

Future research will Focus on incorporating a risk field module in the model to establish pedestrian-vehicle interaction relationships, accurately identify pedestrian position and motion states, and allow safe and comfortable trajectory planning. This approach will contribute to improving the safety and adaptability of AVs in complex road conditions, ensuring the safety of all road users.
Such a method is one of the core parts of parallel planning methodologies \cite{para}. Further methods will be designed and developed to enrich the content of parallel planning, which utilizes the complementarity of online-offline processes and the real-virtual interactive and iterative learning of AVs. These advancements aim to gradually improve the safety and social suitability of AVs as they gain more driving experience.

\ifCLASSOPTIONcaptionsoff
  \newpage
\fi



%



\bibliographystyle{IEEEtran}
\bibliography{bibtex/bib/HAOMO_Sup}

%

\begin{IEEEbiography}[{\includegraphics[width=1in,height=1.25in,clip,keepaspectratio]{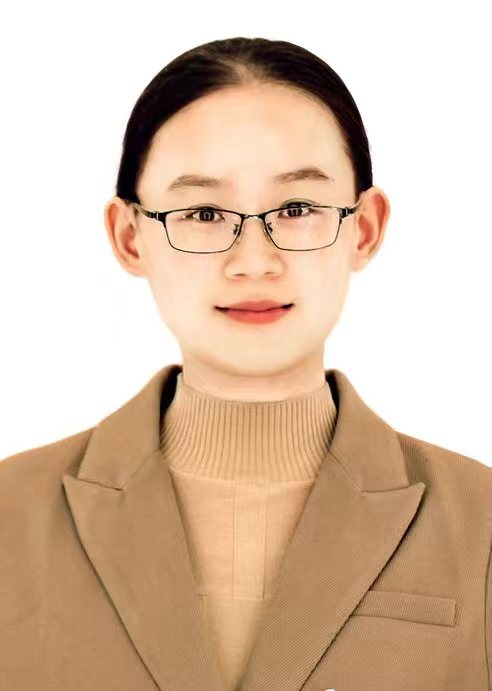}}]{Xiao Wang (Senior Member, IEEE)}
received her bachelor’s degree in network engineering from the Dalian University of Technology, Dalian, China, in 2011, and her Ph.D. degree in social computing from the University of Chinese Academy of Sciences, Beijing, China, in 2016. She is currently a Professor at the School of Artificial Intelligence, Engineering Research Center of Autonomous Unmanned System Technology, Ministry of Education, Anhui University, Hefei, and the President of the Qingdao Academy of Intelligent Industries, Qingdao, China. Her research interests include social transportation, parallel driving, cognitive and embodied intelligence, and multi-agent modeling. 
\end{IEEEbiography}

\begin{IEEEbiography}[{\includegraphics[width=1in,height=1.25in,clip,keepaspectratio]{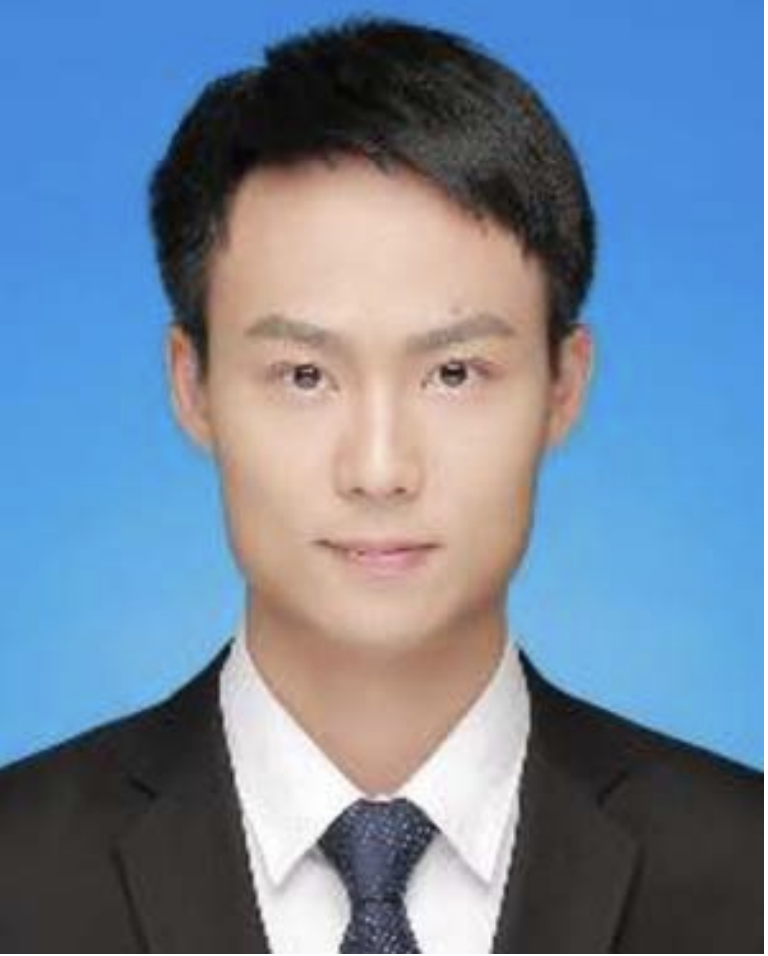}}]{Ke Tang}
received the graduation degree from the Beijing University of Aeronautics and Astronautics and the master’s degree in instrument science and technology from the Beijing University of Aeronautics and Astronautics in 2018. He is currently the Tech Leader of PnC algorithm with Haomo.ai.
\end{IEEEbiography}

\begin{IEEEbiography}[{\includegraphics[width=1in,height=1.25in,clip,keepaspectratio]{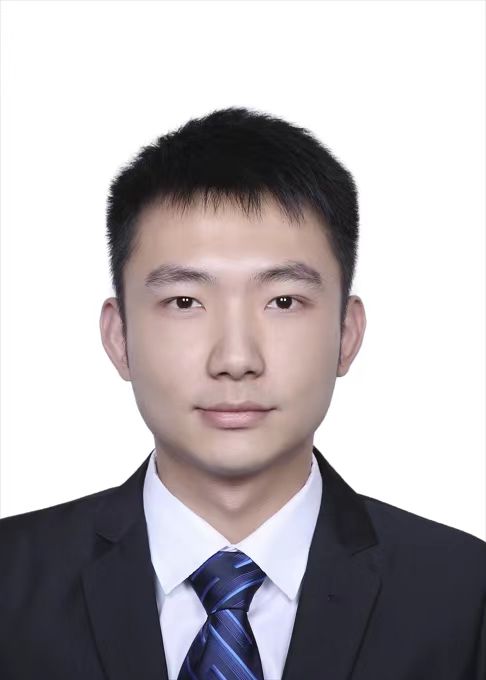}}]{Xingyuan Dai}
received his Ph.D. degree in control theory and control engineering from the Institute of Automation, Chinese Academy of Sciences, Beijing, China, in 2022. He is currently an Assistant Professor with the State Key Laboratory for Management and Control of Complex Systems, Institute of Automation, Chinese Academy of Sciences. His research interest covers intelligent transportation systems, machine learning, and deep learning.
\end{IEEEbiography}

\begin{IEEEbiography}[{\includegraphics[width=1in,height=1.25in,clip,keepaspectratio]{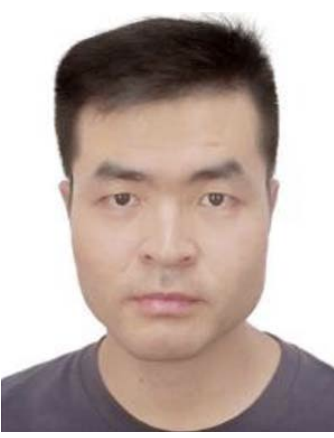}}]{Jintao Xu}
received his Ph.D. degree in control science and engineering from Xi’an Jiaotong University, Xi’an, China, in 2016. He is currently a director of algorithms at Haomo.ai.
\end{IEEEbiography}

\begin{IEEEbiography}[{\includegraphics[width=1in,height=1.25in,clip,keepaspectratio]{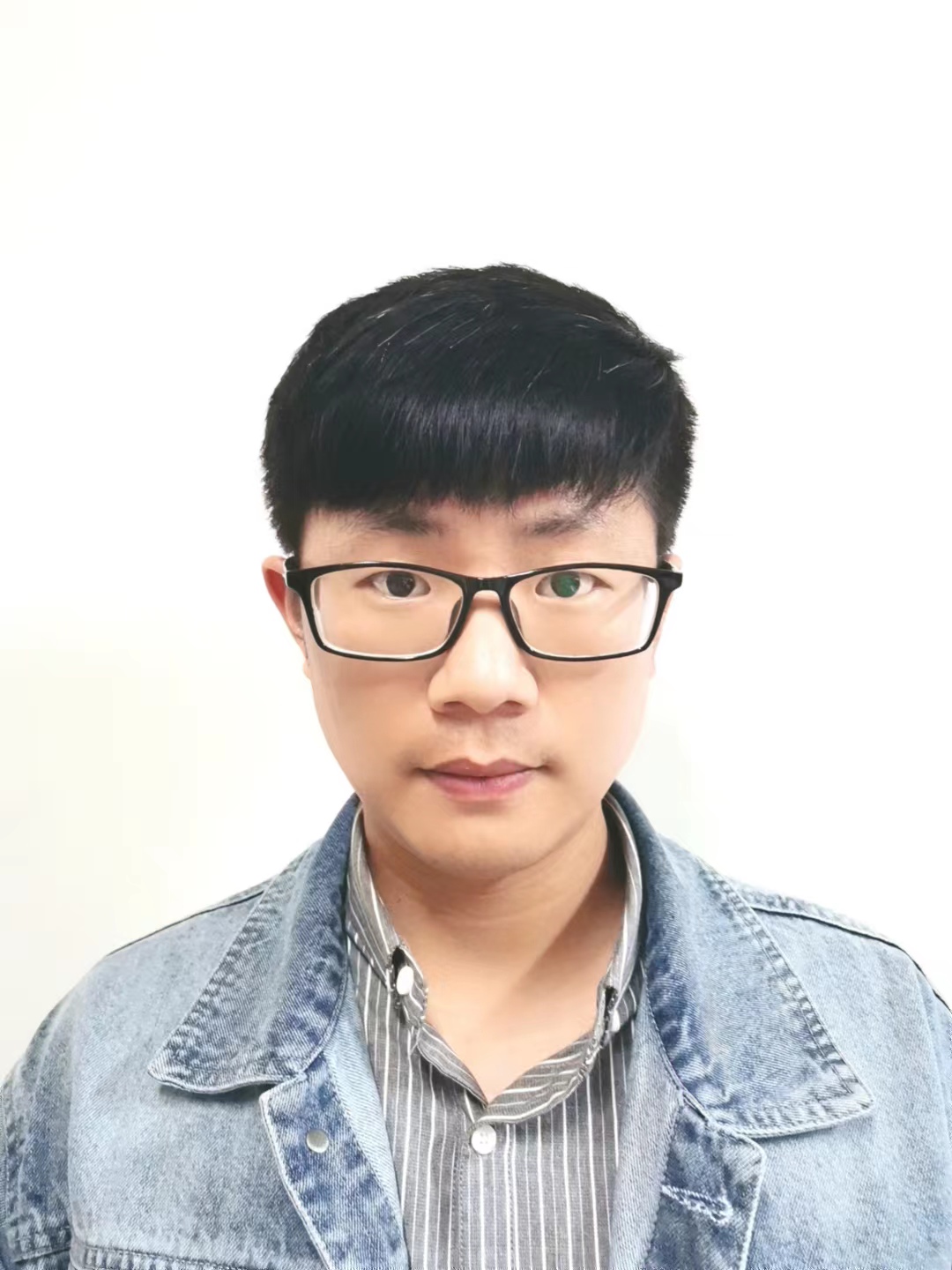}}]{Quancheng Du}
received his B.S. degree from Henan Normal University in 2018 and is currently studying for a Ph.D. degree in the Computer Department of the University of Science and Technology Beijing, China. His main research interests include autonomous driving and trajectory prediction.
\end{IEEEbiography}

\begin{IEEEbiography}[{\includegraphics[width=1in,height=1.25in,clip,keepaspectratio]{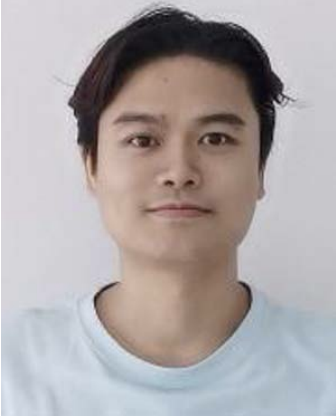}}]{Rui Ai}
received the Ph.D. degree in Pattern Recognition and Intelligent Systems from the Chinese Academy of Sciences in 2013. He is currently the Vice President of Haomo Technology Co., Ltd., and leads the AI Center of Haomo. He is responsible for the research and development of high-level autonomous driving systems, unmanned delivery vehicles, and data intelligence.
\end{IEEEbiography}

\begin{IEEEbiography}[{\includegraphics[width=1in,height=1.25in,clip,keepaspectratio]{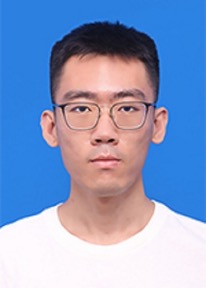}}]{Yuxiao Wang}
received his B.Eng. degree in Automation from Xi'an Jiao Tong University, Xi'an, China, in 2021. He is currently pursuing an MA.Eng. degree in artificial intelligence with the University of Chinese Academy of Sciences, Beijing, China, and the Institute of Automation, Chinese Academy of Sciences, Beijing, China. His research interests include autonomous driving and deep reinforcement learning.
\end{IEEEbiography}

\begin{IEEEbiography}[{\includegraphics[width=1in,height=1.25in,clip,keepaspectratio]{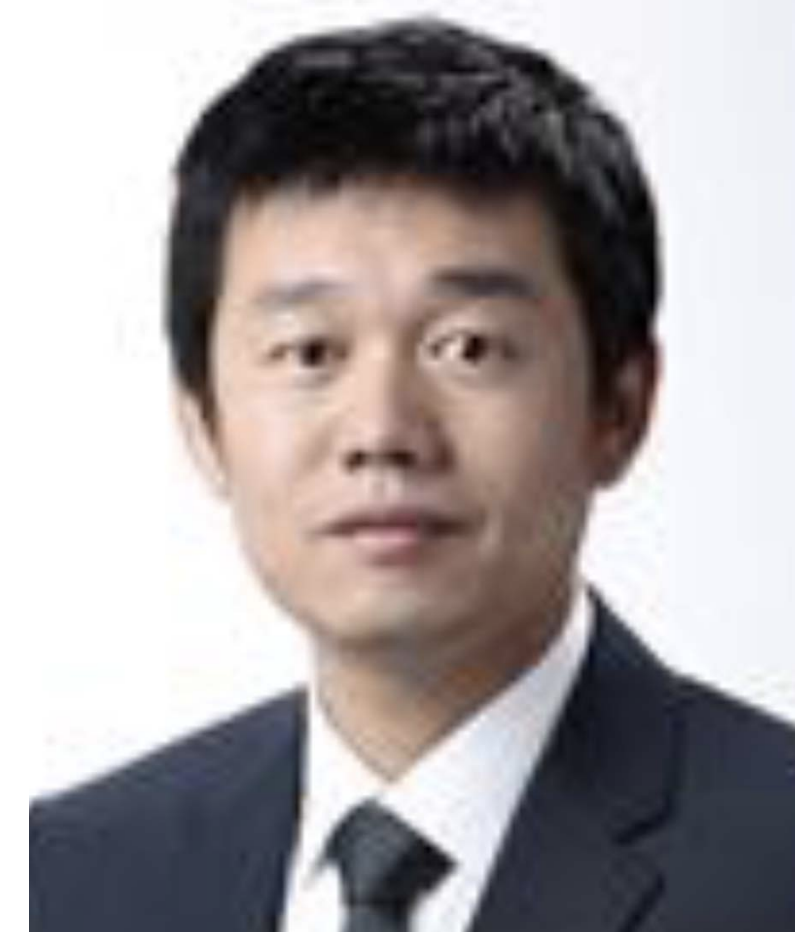}}]{Weihao Gu}
graduated from Beijing Jiaotong University, majoring in computer science. He is the co-founder and CEO of HAOMO.AI, a top-tier startup company in the field of autonomous driving. His current research focuses on autonomous driving, artificial intelligence, and big data. He joined Bai-du in 2003 and held roles including chief architect for MP3 and video search, head of speech recognition technology, deputy general manager of Baidu map, and general manager of the company’s intelligent car division. He led his team to create the first self-driving map in China. He has also developed the first low-cost assisted driving solution and low-cost automatic parking solution in China. He is a pioneer and innovator in the field of autonomous driving in China.
\end{IEEEbiography}

\end{document}